\documentclass[runningheads]{llncs}
\usepackage[T1]{fontenc}
\usepackage{graphicx}
\usepackage{multirow}
\usepackage{colortbl}

\usepackage{amssymb}
\usepackage{amsmath}

\usepackage{amsthm} 

\usepackage{algorithm}
\usepackage[noend]{algorithmic}
\usepackage{bm}

\usepackage{enumitem}

\usepackage{hyperref}
\usepackage{mathtools}

\begin{document}
\title{Reinforcement Learning-based \\Heuristics to Guide Domain-Independent Dynamic Programming}
\titlerunning{RL-based Heuristics to Guide Domain-Independent Dynamic Programming}
%
\author{Minori Narita\inst{1}\orcidID{0000-0003-2808-6056} \and
Ryo Kuroiwa\inst{2}\orcidID{0000-0002-3753-1644} \and\\
J. Christopher Beck\inst{1}\orcidID{0000-0002-4656-8908}}

\authorrunning{M. Narita et al.}
%
\institute{University of Toronto, 5 King’s College Road, Toronto, Ontario, Canada \email{minori.narita@mail.utoronto.ca}, 
\email{jcb@mie.utoronto.ca}\and
National Institute of Informatics, 2-1-2 Hitotsubashi, Chiyoda-ku, \\Tokyo, Japan\\
\email{kuroiwa@nii.ac.jp}}
\maketitle              
\begin{abstract}
Domain-Independent Dynamic Programming (DIDP) is a state-space search paradigm based on dynamic programming for combinatorial optimization. In its current implementation, DIDP guides the search using user-defined dual bounds. Reinforcement learning (RL) is increasingly being applied to combinatorial optimization problems and shares several key structures with DP, being represented by the Bellman equation and state-based transition systems. We propose using reinforcement learning to obtain a heuristic function to guide the search in DIDP. We develop two RL-based guidance approaches: value-based guidance using Deep Q-Networks and policy-based guidance using Proximal Policy Optimization. Our experiments indicate that RL-based guidance significantly outperforms standard DIDP and problem-specific greedy heuristics with the same number of node expansions. Further, despite longer node evaluation times, RL guidance achieves better run-time performance than standard DIDP on three of four benchmark domains.

\end{abstract}

\section{Introduction}
Domain-Independent Dynamic Programming~(DIDP) is a state-space search paradigm based on dynamic programming~(DP) and heuristic state-space search \cite{caasdy,cabs}. Previous work has shown DIDP to be competitive with Constraint Programming~(CP) and Mixed-Integer Programming~(MIP) on a number of benchmark problem classes in combinatorial optimization. 
Current DIDP solvers guide search with an $f$-value computed at each state, where $f(s) = g(s) + h(s)$; $g(s)$ is the path cost to the current state $s$ and $h(s)$ is a heuristic that estimates the cost from $s$ to a base state. In its current implementation, DIDP uses user-defined dual bounds as the $h$-value. However, such dual bounds may not always be very informative and a stronger heuristic guidance could improve solver performance.

Reinforcement learning~(RL) has achieved remarkable performance in fields such as control tasks and games~\cite{a3c,dqn,ppo,alphazero}, and is increasingly being applied to combinatorial optimization~\cite{boisvert2024towards,cappart2021combining,khalil2017learning,nazari2018,jspRL2020}. The goal of the RL agent is to learn an optimal policy for the given task through trial-and-error interactions with an environment described as a Markov Decision Process~(MDP)~\cite{DeepRLSurvey,RLsurvey}. RL shares several key structures with DP, being represented by the Bellman equation and state-based transition systems. Cappart et al.~\cite{cappart2021combining} formulated optimization problems as dynamic programs to bridge an RL model and a CP model, enabling RL-based guidance for variable selection in CP solvers. However, their framework restricts CP formulations to be compatible with the DP model, which can limit performance. In contrast, RL-guided DIDP leverages the alignment between RL and DP models, which may offer a natural integration of RL into exact solvers.

In this paper, we investigate two ways to guide the search in DIDP using RL, value-based and policy-based guidance, and evaluate them on four combinatorial optimization problems. 
The key contributions of this paper are as follows:

\begin{itemize}
    \item We introduce two approaches -- value-based guidance and policy-based guidance -- that effectively direct the search in DIDP;
    \item We demonstrate that an RL model can be systematically mapped from a DIDP model, establishing a basis for automated mapping in future work;
    \item Our experimental evaluation shows that DIDP with RL guidance outperforms DIDP based on node expansions and, to a lesser extent, on run-time.
\end{itemize}

\section{Background}
In this section, we describe the two foundations of our work: Domain-\\Independent Dynamic Programming~(DIDP) and reinforcement learing~(RL).

\subsection{DIDP}
In DIDP, the user defines a DP model in 
the Dynamic Programming Description
Language (DyPDL) and the model is solved by a solver. While different solving approaches are possible, thus far existing solvers are based on heuristic search.

\subsubsection{DyPDL}\label{sec:dypdl}
DyPDL is a solver-independent formalism to define a DP model~\cite{kuroiwa2024journal} represented as a tuple $\langle \mathcal{V}, s_0, \mathcal{T}, \mathcal{B}, \mathcal{C} \rangle$, 
where
$\mathcal{V} = \{v_1, ..., v_n\}$ is the set of state variables, $s_0$ is the target state, $\mathcal{T}$ is the set of transitions, $\mathcal{B}$ is the set of \textit{base cases}, and $\mathcal{C}$ is the set of state constraints. Each state variable $v_i\in \mathcal{V}$ is either an element, a set, or a number, and has a domain $\mathcal{D}_{i}$. A state $s$ is a complete assignment to the state variables, represented by a tuple $\langle d_1, ..., d_n\rangle\in \mathcal{D}$ where $\mathcal{D}$ is the cartesian product of $\mathcal{D}_1$ ... $\mathcal{D}_n$. We denote $s[v_i]$ as the value of the $v_i$ in state $s$. A target state $s_0$ is the initial state in the transition system, i.e., the state for which the optimal value is to be computed. State constraints $\mathcal{C}$ are conditions on state variables that must be satisfied by all valid
states. A base case $\langle \mathcal{C}_B$, \text{\fontfamily{cmss}\selectfont cost}$_B \rangle\in \mathcal{B}$ is a set of conditions $\mathcal{C}_B$ to terminate the transitions and the associated cost function \text{\fontfamily{cmss}\selectfont
cost}$_B$. A state that satisfies $\mathcal{C}\cup \mathcal{C}_B$ is called a base state.
A transition $\tau\in \mathcal{T}$ is a 4-tuple $\langle$\text{\fontfamily{cmss}\selectfont
eff}$_\tau$, \text{\fontfamily{cmss}\selectfont
cost}$_\tau$, \text{\fontfamily{cmss}\selectfont
pre}$_\tau$, \text{\fontfamily{cmss}\selectfont
forced}$_\tau \rangle$. The effect \text{\fontfamily{cmss}\selectfont
eff}$_\tau: \mathcal{D}_{i} \rightarrow \mathcal{D}_{i}$ is a function that maps a value of a state variable $v$ to another value. A state transition returns a successor state $s[\![\tau ]\!]$ by applying $\tau$ to each state variable in $s$, i.e., $s[\![\tau ]\!][v_i]=$ \text{\fontfamily{cmss}\selectfont
eff}$_\tau[v_i](s)$, $\forall v_i\in \mathcal{V}$. 
A numeric cost
\text{\fontfamily{cmss}\selectfont cost}$_\tau(s)$ is associated with each transition $\tau$ from a state $s$.
Preconditions \text{\fontfamily{cmss}\selectfont
pre}$_\tau$ are conditions on state variables, and $\tau$ is \textit{applicable} in a state $s$ only if all preconditions are satisfied, denoted by $s \models$ \text{\fontfamily{cmss}\selectfont
pre}$_\tau$. The flag \text{\fontfamily{cmss}\selectfont
forced}$_\tau\in \{\bot, \top\}$ is a boolean value; if forced transitions are applicable at state $s$, then the first defined one is executed and all other forced and non-forced transitions are ignored.

Let $x=\{x_1,...,x_m\}$ be a sequence of transitions for a DyPDL model. Then, $x$ is a \textit{solution} to the model if the sequence starts from $s_0$ and ends at a base state. 
For minimization problems, the cost of a solution is $\sum_{i=0}^{m-1}$\text{\fontfamily{cmss}\selectfont cost}$_{x_{i+1}}(s_i)+\min_{\{B| B\in \mathcal{B}; s_m\models \mathcal{C}_B\}}$\text{\fontfamily{cmss}\selectfont cost}$_B(s_m)$, where $s_i$ is the state resulting from applying the first $i$ transitions of the solution from $s_0$. 
For maximization, $\min$ is replaced with $\max$.
We can represent a DyPDL model by a recursive equation called a Bellman equation~\cite{bellman1957dynamic}. The Bellman equation $V(s)$ returns the optimal cost starting from state $s$ where $V(s)=\infty$ (or $V(s)=-\infty$ for maximization) if there does not exist a base state reachable from $s$. For the minimization (maximization) problem, $\eta(s)$ is a dual bound function iff $\eta(s) \leq V(s)$ ($\eta(s) \geq V(s)$), $\forall s\in \mathcal{D} \land s\models \mathcal{C}$. 

\subsubsection{State-based heuristic search}
Kuroiwa and Beck~\cite{kuroiwa2024journal} implement seven heuristic search algorithms for DIDP based on the literature. Starting from $s_0$, each algorithm \textit{expands} states, generating a successor state $s'=s[\![\tau ]\!]$ for each transition applicable in $s$. At each successor state, the $f$-value is computed as $f(s')=g(s')+h(s')$, where $g(s')$ is the path cost from $s_0$ to $s'$, and $h(s')$ is the heuristic estimate of the cost from $s'$ to a base state. Given a sequence of transitions $x=\{x_1,...,x_j\}$ to reach a state $s'$~($=s_j$) from $s_0$, the path cost for $s'$ is defined as $g(s') =\sum_{i=0}^{j-1}$\text{\fontfamily{cmss}\selectfont cost}$_{x_{i+1}}(s_i)$. User-defined dual bounds $\eta(s)$ and state constraints $\mathcal{C}$ are used for pruning. 
Let $\bar{\zeta}$ be the the primal bound. Then, we can prune the node $s$ if $g(s) + \eta(s)\geq \bar{\zeta}$~(for maximization, $g(s) + \eta(s)\leq \bar{\zeta}$). By default, $h(s') = \eta(s')$. For formal details, see Kuroiwa and Beck~\cite{kuroiwa2024journal}. 

In Section~\ref{sec:experiments}, we experiment with three search algorithms: complete anytime beam search (CABS), anytime column progressive search (ACPS), and anytime pack progressive search (APPS). Algorithm details appear in Appendix A. 

\subsection{Reinforcement Learning}
Reinforcement learning~(RL)~\cite{sutton1998reinforcement} is a framework for learning to achieve a goal from interaction with the environment. In RL, the agent operates based on a Markov Decision Process~(MDP)~\cite{mdp}, defined as a 4-tuple $\langle \mathrm{S, A, T, R}\rangle$, where $\mathrm{S}$ is a set of states, $\mathrm{A}$ is a set of actions, $\mathrm{T}: \mathrm{S}\times \mathrm{A}\rightarrow \mathrm{S}$ is the transition function, and $\mathrm{R}: \mathrm{S}\times \mathrm{A}\rightarrow \mathbb{R}$ is the reward function. For simplicity, we use the notation $s$ to represent a state in the MDP as well as a state in the DP. The initial state $s_0$ is sampled from an initial state distribution $\rho_0$: $\mathbb{R}\rightarrow \mathrm{S}$. At each time-step, the agent performs an action $a\in \mathrm{A}$ at current state $s$, which brings the agent to the next state $s'$ and gives a reward $r=\mathrm{R}(s, a)$. We assume a deterministic transition function, so $\mathrm{T}(s, a)$ returns the next state $s'$. An episode terminates when the agent reaches a terminal state. The goal of the RL agent is to learn an optimal policy $\pi$, that maximizes the expected total reward $\sum_{t=0}^{\infty} \gamma^t r_{t}$, given the initial state distribution $\rho_0$, where $\gamma$~is a discount factor~($0\leq \gamma \leq 1$) and $t$ is a time-step. A policy $\pi: \mathrm{S}\times \mathrm{A}\rightarrow [0, 1]$ is a conditional distribution over actions given the state, indicating the likelihood of the agent choosing an action. Policy-based methods like TRPO~\cite{trpo} and PPO~\cite{ppo} directly optimize the policy through policy gradient methods. Value-based methods such as DQN~\cite{dqn} learn a value function from exploration. For instance, DQN learns an estimated Q-value function $Q^{\pi}(s, a)$, the expected return for selecting action~$a$ at state~$s$ if the agent follows policy~$\pi$ afterwards.

RL is increasingly being applied to solve combinatorial optimization problems. Early approaches used recurrent neural networks to learn constructive heuristics for generating solutions~\cite{bello2016neural}. However, graph neural networks~(GNNs) have become more prevalent~\cite{khalil2017learning,kool2018attention,nazari2018} as they are size-agnostic and permutation-invariant. A limitation of end-to-end RL methods is the challenge of managing constraints, as well as the lack of a systematic way to improve the obtained solutions, unlike exact methods such as CP and MIP~\cite{cappart2023combinatorial,cappart2021combining}.

Significant progress has recently been made in combining search with learned heuristics to address these limitations~\cite{cappart2021combining,fu2021generalize,kool2022deep,sun2023difusco}. Kool et al.~\cite{kool2022deep} formulated routing problems as a DP problem and guided the search using a learned heat map of edge weights. Cappart et al.~\cite{cappart2021combining} formulated optimization problems as DP models to bridge an RL model and a CP model, enabling RL-based guidance for variable selection in CP solvers. 

Guiding search using deep reinforcement learning has also been explored in AI planning. Orseau et al.~\cite{orseau2018,orseau2021policy} proposed learning Q-value functions and policies to weight nodes in best-first search to solve two-dimensional single-agent problems like Sokoban. DeepCubeA~\cite{deepcubeA} tackled the Rubik's Cube by learning a heuristic function through approximated value iteration and applying it in batch-weighted A* search. Lastly, Gehring et al.~\cite{gehring2022reinforcement} leveraged domain-independent heuristic functions as dense reward generators to train RL agents, and used the RL-based heuristics to improve search efficiency for classical planning problems.

\section{RL-based search guidance for DIDP}

Given the similarities in the state-based formulations of DIDP and RL, it is natural to investigate the guidance of search in DIDP based on a heuristic function learned with RL. Each component of an MDP can be systematically derived from the corresponding component in a DIDP model. For instance, the set of states in the DIDP model matches precisely with the state space in the MDP and transitions in the DIDP model have a 1-to-1 mapping to actions in the MDP. An RL agent trained on the mapped MDP can then serve as a heuristic function for computing $f$-values in the DIDP search. 

\begin{figure}[tb]
    \begin{center}
        \includegraphics[width=0.82\linewidth]{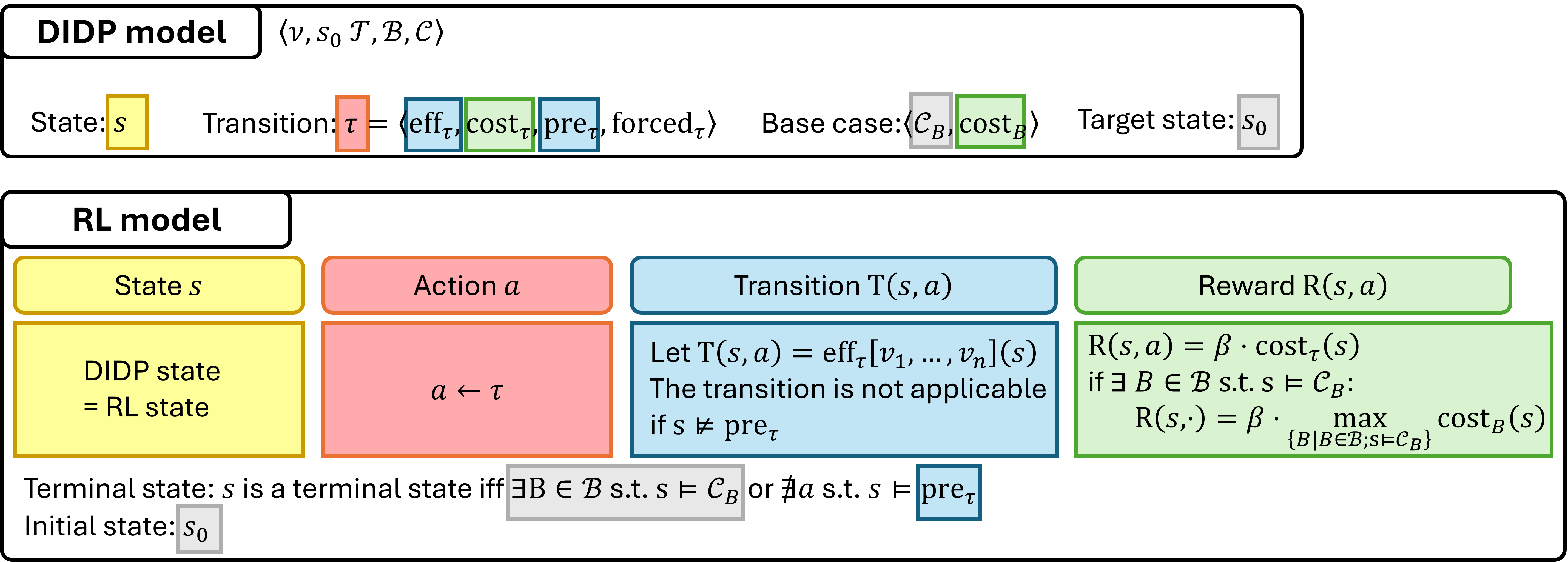}
    \end{center}
    \caption{Mapping from a DIDP model to an RL model for maximization problems. State constraints $\mathcal{C}$ and \text{\fontfamily{cmss}\selectfont
forced}$_\tau$ are not mapped to the RL model.} 
    \label{fig:dp_rl_mapping}
\end{figure}

In this paper, we develop a scheme to convert a DIDP model to an RL model for a given combinatorial optimization problem. Figure~\ref{fig:dp_rl_mapping} provides an overview of how an RL model is mapped from a DIDP model. Each component in the MDP is derived from the DIDP model as follows:

\begin{itemize}[label=$\bullet$]
    \item \texttt{State} $s\in \mathrm{S}$: The same state space as in the DIDP model.
    
    \item \texttt{Action} $a\in \mathrm{A}$: 1-to-1 mapping from transition $\tau\in \mathcal{T}$, i.e., $\tau \mapsto a$.
    
    \item \texttt{Transition function} $\mathrm{T}(s, a)$: mapped from \text{\fontfamily{cmss}\selectfont
pre}$_\tau$ and \text{\fontfamily{cmss}\selectfont
eff}$_\tau$ in each $\tau\in\mathcal{T}$. For a state $s$, the next state reached by taking an action $a$ is obtained by applying \text{\fontfamily{cmss}\selectfont
eff}$_\tau$ to each state variable in $s$, i.e., $\mathrm{T}(s, a)=$ \text{\fontfamily{cmss}\selectfont
eff}$_\tau[v_1, ..., v_n](s)$. The transition is not applicable if the preconditions are not met, i.e., $s\nvDash$ \text{\fontfamily{cmss}\selectfont
pre}$_\tau$. Thus, the mapping includes the masking of non-applicable actions as is common in RL models with invalid actions~\cite{cappart2021combining,huang2020closer,alphazero}.

    \item \texttt{Reward function} $\textrm{R} (s, a)$ corresponds to the transition cost incurred by applying $\tau$ at state $s$, \text{\fontfamily{cmss}\selectfont
cost}$_\tau(s)$. To improve the stability of RL training, the reward is scaled by a hyperparameter $\beta$.
For a minimization problem, the reward has to be negated to make the RL task a maximization problem,
 i.e., $\textrm{R} (s, a)=-\beta\cdot$ \text{\fontfamily{cmss}\selectfont
cost}$_\tau(s)$. If $s$ is the base state, \text{\fontfamily{cmss}\selectfont
cost}$_B$ is applied instead.
\end{itemize}

A state $s$ is a terminal state if $s$ satisfies at least one base case, i.e., $\exists \mathrm{B}\in \mathcal{B}$ s.t. $s\models\mathcal{C}_\mathrm{B}$, or there is no action that satisfies preconditions, i.e., $\nexists$ $a$ s.t. $s\models$\text{\fontfamily{cmss}\selectfont
pre}$_\tau$ where $a$ is mapped from $\tau$. The initial state in RL is the target state in the DIDP model $s_0$. The state constraints $\mathcal{C}$ are not mapped to the RL model.\footnote{While state constraints $\mathcal{C}$ can be mapped to action masking, the implications of action masking in RL remain underexplored~\cite{huang2020closer}. Thus, the investigation of integrating $\mathcal{C}$ into the MDP is left for future work.}

Including domain knowledge and introducing auxiliary rewards are often crucial for successful RL training~\cite{HRL_aux_reward,alphago}. 
In this paper, we focus on leveraging the structural relationships between DIDP and RL models to systematically build RL models for search guidance. 
Future work will examine automating this mapping and incorporating problem-specific structures into the mapping process.

\begin{figure}[tb]
    \begin{center}
        \includegraphics[width=0.98\linewidth]{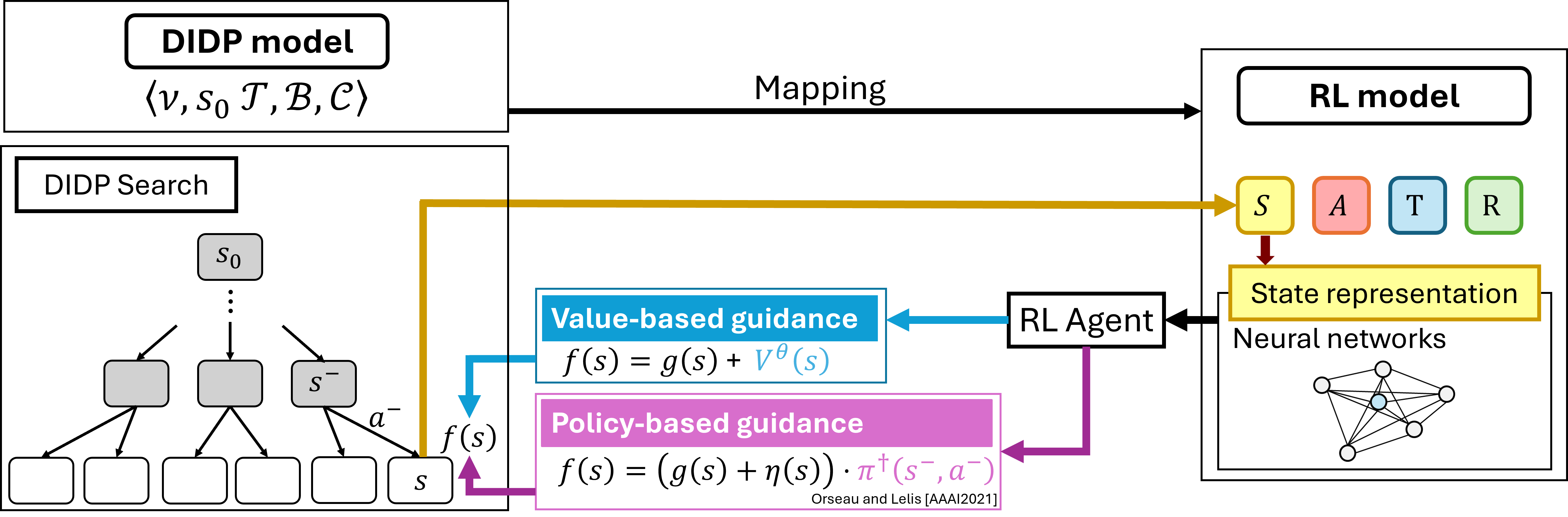}
    \end{center}
    \caption{Value-based guidance and policy-based guidance for DIDP. The equations for computing $f$-values in the figure are for maximization problems.} 
    \label{fig:rl_guidance}
\end{figure}

\subsection{Value-based guidance}\label{sec:value-based}

The value-based guidance approach directly uses the value function approximated by neural network parameters $\theta$, $V^{\theta}(s)$, as a heuristic function, where $h(s)=V^{\theta}(s)$. $V^{\theta}(s)$ is an estimated total reward from $s$ to a terminal state, which aligns with the definition of $h(s)$, the estimated total cost from $s$ to a base state. For minimization problems, the value function is negated, i.e., $h(s)=-V^{\theta}(s)$. Figure~\ref{fig:rl_guidance} shows the overview of this approach. First, the MDP is derived from the DIDP model, as described above. Then, the RL agent is trained in the mapped MDP with a value-based RL algorithm~(e.g., DQN). 
During search, every time a node is expanded, the $f$-value is calculated for all its successor states $s$ by $f(s)=g(s)+V^{\theta}(s)$; the $f$-values are then used to determine the priority of the state in the DIDP framework. 

Details of the computation for each successor node $s$ are as follows. First, path cost $g(s)$ is calculated by $g(s) = g(s^-) + $\text{\fontfamily{cmss}\selectfont cost}$_\tau(s^-)$, where $s^-$ is the parent node of $s$ and $\tau$ is a transition that transforms $s^-$ to $s$.
To align the scale of $g(s)$ with that of $V^{\theta}(s)$, the scaling factor $\beta$ is used; thus, $g(s)=g(s^-)+ \beta \cdot$ \text{\fontfamily{cmss}\selectfont cost}$_\tau(s^-)$. The heuristic value is computed by calling a neural network prediction, i.e., $V^{\theta}(s)=\mathcal{F}(s;\theta)$. We use DQN as a value-based RL algorithm, so the output is the Q-values for each action $a$ for the input state $s$. $V^{\theta}(s)$ is obtained by $V^{\theta}(s)=\max_{a\in A'} Q^{\theta}(s, a)$, where $A'$ is the set of all applicable actions in $s$. 


\subsection{Policy-based guidance}\label{sec:policy-based}
The policy-based guidance approach uses the same MDP as value-based guidance, but is trained with a policy-based RL algorithm, such as PPO~\cite{ppo}. The algorithm learns a policy $\pi(s, a)$, a probability distribution over actions for a given state. $\pi(s,a)$ is then used to weight the original $f$-value to prioritize the expansion of the nodes that are deemed promising by the policy. 

Details of the computation at each successor node generation are as follows. First, path cost $g(s)$ is calculated by $g(s) = g(s^-) + $\text{\fontfamily{cmss}\selectfont cost}$_\tau(s^-)$. Unlike in value-based guidance, there is no need to scale the path cost $g(s)$, as the policy has a fixed scale of $[0, 1]$ and is only used to weight the original $f$-values. Then, the policy $\pi(s^-,a^-)$ is obtained by calling the neural network and obtain the $a^-$th output, i.e., $\pi(s^-,a^-)=\mathcal{F}(s^-;\theta)[a^-]$.\footnote{In our implementation, the neural network is called only once when $s^-$ is expanded. Each element of $\pi(s^-)$ is assigned to the corresponding successor through indexing.} Our approach uses the accumulated probabilities up to state $s$ from the root node, i.e., $\pi^{\dag}(s^-, a^-)=\pi(s_0, a_0)\pi(s_1, a_1)...\pi(s^-, a^-)$, to take all the previous decisions up to $s$ into consideration~\cite{orseau2021policy}. Therefore, the $f$-value is computed as $f(s)=(g(s)+\eta (s))\cdot\pi^{\dag}(s^-, a^-)$, where $\eta(s)$ is the dual bound at $s$. A promising action with a high probability in the policy will have a higher $f$-value~(for maximization), thereby making the corresponding state more likely to be expanded next. For minimization problems, we divide $g(s)+\eta(s)$ by $\pi^{\dag}(s^-, a^-)$ instead, i.e., $f(s)=(g(s)+\eta (s))/ \pi^{\dag}(s^-, a^-)$, so that promising actions yield lower $f$-values.

\subsection{State Representation}\label{sec:neuralnet}
The state representation needs to be able to handle instances of different sizes and to be invariant to input permutations~\cite{cappart2021combining}. Hence, we used a graph attention network~(GAT)~\cite{gat} as a state representation for routing problems to leverage the natural graph structure of these domains, and a set transformer~\cite{settransformer} or Deep Sets~\cite{deepsets} for packing problems. The details of the neural network architecture for each problem domain used in this paper are provided in Appendix~\ref{appendix:architecture}. The embedding obtained by the neural network can then be used as an input to a fully-connected network. For DQN, the network outputs Q-values for each action $a$ for state $s$, so the output layer is of size $|A|$. For PPO, two separate networks for the actor and the critic are used. The critic network outputs a single value, representing the estimated value of the state, while the actor network applies a softmax activation function to its final layer to output action probabilities $\pi(s,\cdot)$. The outputs are processed as described in Sections~\ref{sec:value-based} and \ref{sec:policy-based}.

\section{Experiments}\label{sec:experiments}
We evaluated our methods on four combinatorial optimization problems: Traveling Salesperson Problem (TSP), TSP with Time Windows (TSPTW), 0-1 Knapsack, and Portfolio Optimization.\footnote{All code are available at \url{https://github.com/minori5214/rl-guided-didp}.} To assess the quality of RL guidance, the solution quality per node expansion was evaluated for different guidance methods.
Our two RL-based guidance methods~($h$=DQN and $\pi$=PPO) were compared with dual-bound guidance~(default DIDP implementation), uniform cost search~(i.e., $h=0$ for all states), and greedy heuristic guidance. 

The greedy heuristic-based $h$-value at a state is equal to the cost of the path to a base case found by rolling-out the greedy heuristic from that state. 
Greedy heuristics exploit domain-specific knowledge from outside the DP model, and thus serve as a baseline to evaluate how RL guidance competes with hand-crafted heuristics. The definitions of the greedy heuristics for each problem domain and further details are in Appendix~\ref{appendix:greedy}.

We also evaluated the solution quality after a one hour run-time to compare the performance of our approach with baseline methods. The results include performance from MIP~(Gurobi), pure CP~(CPLEX CP Optimizer), RL-guided CP (BaB-DQN and RBS-PPO~\cite{cappart2021combining}), and sampling-based heuristic methods~(dual-bound, greedy, DQN, PPO). Sampling is done by applying a softmax function to the $h$-values of all successor states to get action probabilities and choosing the next state probabilistically. The number of samples is set to 1280, following Kool et al.~\cite{kool2018attention}. The memory limit for each approach is set to 8 GB.

\vspace{0.2cm}
\noindent \textbf{Evaluation metric: }At a given node expansion limit $l$, the gap is calculated by $gap = |x(i, m, l)-best(i)|/best(i)\times 100$, where $x(i, m, l)$ is the solution cost of method $m$ for instance $i$ up to the node expansion limit $l$, and $best(i)$ is the best known solution for $i$. The best known solution includes results from all the approaches, including baselines, within the time limit. If no feasible solution is found within the given node expansion limit, a fixed value of 100~[\%] is used. 

\vspace{0.2cm}
\noindent \textbf{Training Process: }Although the network architectures are size-agnostic, we trained DQN and PPO for each problem size and domain, as examining the scalability of neural networks 
is not our main focus. Training begins with randomly generating a problem instance from a fixed distribution using an instance generator. The RL agent then explores the instance following the current policy $\pi$ until the agent reaches the terminal state with the experiences stored in the replay buffer. The network parameters $\theta$ are updated using the batched experiences sampled from the replay buffer. The training time is limited to 72 hours. Details of the network architectures and hyperparameters are in Appendix~\ref{appendix:architecture}.

\subsection{Problem Domains}
To evaluate our approach, we chose routing problems~(TSP and TSPTW) and packing problems (0-1 Knapsack and Portfolio Optimization). 

\subsubsection{TSP}
In the Traveling Salesperson Problem~(TSP)~\cite{tsp}, a set of customers $N=\{0,...,n\}$ is given, and a solution is a tour starting from the depot~($i=0$) and returning to the depot, visiting each customer exactly once. Visiting customer $j$ from $i$ incurs the travel time $c_{ij}\geq0$. TSP instances are generated by removing time window constraints from the TSPTW instances used below.

\vspace{0.2cm}
\noindent \textbf{DIDP model: }For TSP, a state is a tuple of variables $\langle U, i\rangle$ where $U$ is the set of unvisited customers and $i$ is the current location. In this model, one customer is visited at each transition. The minimum possible travel time to visit customer $j$ is $c_j^{\text{in}} = \min_{k\in N\setminus \{j\}} c_{kj}$, and the minimum travel time from $j$ is $c_j^{\text{out}} = \min_{k\in N\setminus \{j\}} c_{jk}$. The DIDP model is represented by the following Bellman equation, adapted from the TSPTW model defined below.
\begin{align}
    &\textrm{compute } V(N\setminus \{0\}, 0)\label{eq:tsp0}\\
    &V(U, i) = \left\{
    \begin{array}{ll}
    c_{i0} \hspace{3.6cm} \textrm{ if }U=\emptyset\\
    \min_{j\in U} c_{ij} + V(U\setminus \{j\}, j) \textrm{  if } U\neq \emptyset
    \end{array}
    \right.\label{eq:tsp1}\\
    &V(U, i)\geq \max \left\{\sum\nolimits_{j\in U\cup \{0\}} c_j^{\text{in}}, \sum\nolimits_{j\in U\cup \{i\}} c_j^{\text{out}} \right\}\label{eq:tsp3}
\end{align}

Expression~(\ref{eq:tsp0}) declares that the optimal cost is the cost to visit all customers~($U=N\setminus \{0\}$) starting from the depot~($i=0$).
The second line of Eq.~(\ref{eq:tsp1}) corresponds to visiting customer $j$ from $i$; then, $j$ is removed from $U$ and the current location $i$ is updated to $j$. The first line of Eq.~(\ref{eq:tsp1}) is the base case, where all customers are visited~($U=\emptyset$) and the recursion ends.
Eq.~(\ref{eq:tsp3}) represents two dual bounds. 

\noindent \textbf{RL model: } The MDP for this DIDP model is defined as follows:
\begin{itemize}
    \item[] \texttt{State} $s$: $\langle U, i\rangle$
    \item[] \texttt{Action} $a = j\in U$
    \item[] \texttt{Transition function} $\text{T}(s, a)$: $\text{T}(\langle U, i\rangle, j)=\langle U\setminus {j}, j\rangle$
    \item[] \texttt{Reward function} $\text{R}(s, a)$: $\text{R}(\langle U, i\rangle, j)=\beta\cdot(-c_{ij})$
\end{itemize}

The reward function is the negative distance between the current location $i$ and the next customer $j$. The scaling factor is $\beta=0.001$. Dual bounds are not used in the MDP.

\subsubsection{TSPTW} In TSP with Time Windows (TSPTW)~\cite{tsptw}, the visit to customer $i$ must be within a time window $[a_i,b_i]$. If customer $i$ is visited before $a_i$, the salesperson has to wait until $a_i$. The instances were generated in the same way as Cappart et al.~\cite{cappart2021combining}, but the maximal time window length allowed is set $W=100$, and the maximal gap between two consecutive time windows is set $G=1000$ to make the instances more challenging. 

\vspace{0.2cm}
\noindent \textbf{DIDP model: } A state is a tuple $\langle U, i, t\rangle$ where $t$ is the current time. 
The set of customers that can be visited next is $U'=\{j\in U\mid t+c_{ij}\leq b_j\}$. The DIDP model is represented by the following Bellman equation~\cite{kuroiwa2024journal}:
\begin{align}
    &\textrm{compute } V(N\setminus \{0\}, 0, 0)\label{eq:tsptw0}\\
    &V(U, i, t) = \left\{
    \begin{array}{ll}
    c_{i0} \hspace{6.2cm} \textrm{ if }U=\emptyset\\
    \min_{j\in U'} c_{ij} + V(U\setminus \{j\}, j, \max (t+c_{ij}, a_j)) \textrm{  if } U\neq \emptyset
    \end{array}
    \right.\label{eq:tsptw1}\\
    &V(U, i, t) = \infty \hspace{4.8cm} \textrm{if }\exists j\in U, t+c^*_{ij}>b_j\label{eq:tsptw2}\\
    &V(U, i, t)\leq V(U, i, t') \hspace{3.7cm}  \textrm{if }t\leq t'\label{eq:tsptw3}\\
    &V(U, i, t)\geq \max \left\{\sum\nolimits_{j\in U\cup \{0\}} c_j^{\text{in}}, \sum\nolimits_{j\in U\cup \{i\}}
    c_j^{\text{out}} \right\}\label{eq:tsptw4}
\end{align}

In the second line of Eq.~(\ref{eq:tsptw1}), time $t$ is updated to $\max (t+c_{ij}, a_j)$. Eq.~(\ref{eq:tsptw2}) is a state constraint that sets the value of a state to be infinity if there exists a customer $j$ that cannot be visited by the deadline $b_j$ even if we take the shortest path with distance $c^*_{ij}$. Inequality~(\ref{eq:tsptw3}) is a dominance relationship; if other state variables are the same in two states, then a state having smaller $t$ always leads to a better solution. Eq.~(\ref{eq:tsptw4}) represents two dual bounds.

\noindent \textbf{RL model: } The MDP for this DIDP model is defined as follows:
\begin{itemize}
    \item[] \texttt{State} $s$: $\langle U, i, t\rangle$
    \item[] \texttt{Action} $a = j\in U$
    \item[] \texttt{Transition function} $\text{T}(s, a)$: $\text{T}(\langle U, i, t\rangle, j)=\langle U\setminus {j}, j, \max(t+c_{ij}, a_j)\rangle$
    \item[] \texttt{Reward function} $\text{R}(s, a)$: $\text{R}(\langle U, i, t\rangle, j)=\beta\cdot(|\text{UB}_{cost}|-c_{ij})$
\end{itemize}

$\text{UB}_{cost}$ is a strict upper bound on the reward of any solution for this problem domain to ensure that the RL agent has the incentive to find feasible solutions first and then to find the best ones. $\text{UB}_{cost}$ is not included in the mapping in Figure~\ref{fig:dp_rl_mapping}, but this reward structure is originally introduced in Cappart et al.~\cite{cappart2021combining} and helps improve the RL training. The scaling factor is set to $\beta = 0.001$.

\subsubsection{0-1 Knapsack}
In the 0-1 Knapsack Problem~\cite{knapsack}, a set of items $N=\{0, ..., n-1\}$ with weights $w_i$ and profits $p_i$ for $i\in N$ and a knapsack with budget $B$ are given. The objective is to maximize the total profit of the items in the knapsack. The items are sorted in descending order of the profit ratio~($p_i/w_i$). The instance distribution is taken from the ``Hard'' instances in Cappart et al.~\cite{cappart2021combining}, where profits and weights are strongly correlated.

\vspace{0.2cm}
\noindent \textbf{DIDP model: }A DIDP state is a tuple $\langle x, i\rangle$, where $x$ is the current total weight and $i$ represents the current item index. The DIDP model is based on Kuroiwa and Beck~\cite{kuroiwa2024journal} with the remaining budget replaced by the current total weight $x$:

\begin{align}
    &\textrm{compute } V(0, 0)\label{eq:kp0}\\
    &V(x, i) = \left\{
    \begin{array}{ll}
    \max \{p_i + V(x+w_i, i+1), V(x, i+1)\}\\ \hspace{3cm}\textrm{ if } i<n \land x+w_i\leq B\\
    V(x, i+1)\hspace{1.3cm}\textrm{ if } i<n \land x+w_i>B\\
    0\hspace{2.8cm}\textrm{ otherwise.}
    \end{array}
    \right.\label{eq:kp1}\\
    &V(x, i)\leq \min \left\{\sum_{j=i}^{n-1}p_j, \max_{j\in \{i..n-1\}} \left(\frac{p_j}{w_j}\right)\cdot (B-x)\right\}.\label{eq:knapsack_dual}
\end{align}

Expression~(\ref{eq:kp0}) declares that the optimal cost is the cost to consider all items starting from the first item~($i=0$) with the current total weight $x=0$.
The first line of Eq.~(\ref{eq:kp1}) corresponds to considering item $i$; if $i$ is taken~(the first term), then $x$ is updated to $x+w_i$ and the item index is updated to $i+1$. If $i$ is not taken~(the second term), $x$ remains the same. The second line of Eq.~(\ref{eq:kp1}) indicates that $i$ cannot be taken if doing so exceeds budget $B$. The third line is the base case; when all items are visited~($i\geq n$), then the recursion terminates. Eq.~(\ref{eq:knapsack_dual}) represents two dual bounds. 

\vspace{0.2cm}
\noindent \textbf{RL model: } The MDP for the RL agent is as follows:
\begin{itemize}
    \item[] \texttt{State} $s$: $\langle x, i\rangle$
    \item[] \texttt{Action} $a\in \{0, 1\}$: whether to take the item $i$ or not.
    \item[] \texttt{Transition function} $\text{T}(s, a)$: $\text{T}(\langle x, i\rangle, a)=\langle aw_i+x, i+1\rangle$
    \item[] \texttt{Reward function} $\text{R}(s, a)$: $\text{R}(\langle x, i\rangle, a)=\beta(ap_i)$
\end{itemize}

The RL state is same as the DP state, and the action set is binary: 0 indicates the item is not selected, while 1 means it is selected. The transition function matches the effect of a transition in the DIDP model. The state variable $x$ is updated to $x+aw_i$, i.e., $w_i$ is added to $x$ if the item is selected ($a=1$). Also, the item index $i$ is incremented by 1. The reward function corresponds to $p_i$, the profit of item $i$. The scaling factor is set to $\beta=0.0001$.

\subsubsection{Portfolio Optimization}
In the 4-moments portfolio optimization problem~\cite{portfolio}, a set of investments $N=\{0, ..., n-1\}$, each with a specific cost ($w_i$), expected return ($\mu_i$), standard deviation ($\sigma_i$), skewness ($\gamma_i$), and kurtosis ($\kappa_i$), and the budget $B$ are given. The goal is to find a portfolio with a maximum return as specified by the objective function~(Eq.~(\ref{eq:g_portfolio})). Each financial characteristic is weighted ($\lambda_1$, $\lambda_2$, $\lambda_3$, and $\lambda_4$). The instance distribution is taken from Cappart et al.~\cite{cappart2021combining}. 

\noindent \textbf{DIDP model: }A state is a tuple $\langle x, i, Y\rangle$, where $x$ is the current total weight, $i$ is the current item index, and $Y$ is a set of investments up to $i$, $\{0, ..., i-1\}$. The objective function value up to item $i$ is defined as follows:

\begin{equation}
    \nu(Y)=\lambda_1 \sum_{j\in Y}\mu_j-\lambda_2\sqrt[2]{\sum_{j\in Y}\sigma^2_j}+\lambda_3\sqrt[3]{\sum_{j\in Y}\gamma^3_h}-\lambda_4\sqrt[4]{\sum_{j\in Y}\kappa^4_j}.\label{eq:g_portfolio}
\end{equation}

The transition cost is the difference between the objective value of the current state~($\nu(Y)$) and that of the successor state, i.e., $\nu(Y\cup \{i+1\})-\nu(Y)$. The DIDP model is expressed as follows:

\begin{align}
    &\textrm{compute } V(0, 0, \emptyset)\\
    &V(x, i, Y) = \left\{
    \begin{array}{ll}
    \max (\nu(Y\cup \{i\})- \nu(Y) +V(x+w_i, i+1, Y\cup \{i\}), \\
    \hspace{1.2cm} V(x, i+1, Y)) \hspace{2cm}
    \textrm{ if } i<n \land x+w_i\leq B\\
    V(x, i+1, Y)\hspace{3.35cm}\textrm{ if } i<n \land x+w_i>B\\
    0\hspace{5.2cm}\textrm{ otherwise.}\\
    \end{array}
    \right.\label{eq:portfolio1}\\
    &V(x, i, Y)\leq \min\left(\lambda_1 \sum_{j=i}^{n-1}\mu_j+\lambda_3\sqrt[3]{\sum_{j=i}^{n-1}\gamma^3_j}, \max_{j\in \overline{Y}} K_j\cdot (B-x)\right)\label{eq:portfolio_dual}
\end{align}

\noindent
where $K_j=\left(\frac{\lambda_1 \mu_j + \lambda_3 \sqrt[3] {\gamma^3_j}}{w_j}\right)$ and $\overline{Y}=\{i..n-1\}$. In the first line of Eq.~(\ref{eq:portfolio1}), if item $i$ is taken, then $Y$ is updated to $Y\cup \{i\}$. The second and third lines of Eq.~(\ref{eq:portfolio1}) are the same as Eq.~(\ref{eq:kp1}) in 0-1 Knapsack except that a state includes $Y$ as well. Eq.~(\ref{eq:portfolio_dual}) represents two dual bounds. Proofs for the dual bounds are shown in Appendix~\ref{appendix:portfolio_duals}.

\vspace{0.2cm}
\noindent \textbf{RL model: } The MDP for the RL agent is as follows:

\begin{itemize}
    \item[] \texttt{State} $s$: $\langle x, i, Y\rangle$
    \item[] \texttt{Action}: $a\in \{0, 1\}$
    \item[] \texttt{Transition function} $\text{T}(s, a)$: 
    \item[] $\hspace{2.0cm}T(\langle x, i, Y\rangle, 1) = \langle w_i + x, i + 1, Y \cup \{i\} \rangle$
    \item[] $\hspace{2.0cm}T(\langle x, i, Y\rangle, 0) = \langle x, i + 1, Y \rangle$
    \item[] \texttt{Reward function} $\text{R}(s, a)$: 
    $R(\langle x, i, Y\rangle, a)=a \cdot (\nu(Y \cup \{i\}) - \nu(Y))$
\end{itemize}

In the transition function, $Y$ is updated to $Y\cup \{i\}$ if $i$ is taken.
The scaling factor is set to $\beta=0.0001$.

\section{Results}

\begin{figure}[t!]
    \begin{center}
        \includegraphics[width=0.98\linewidth]{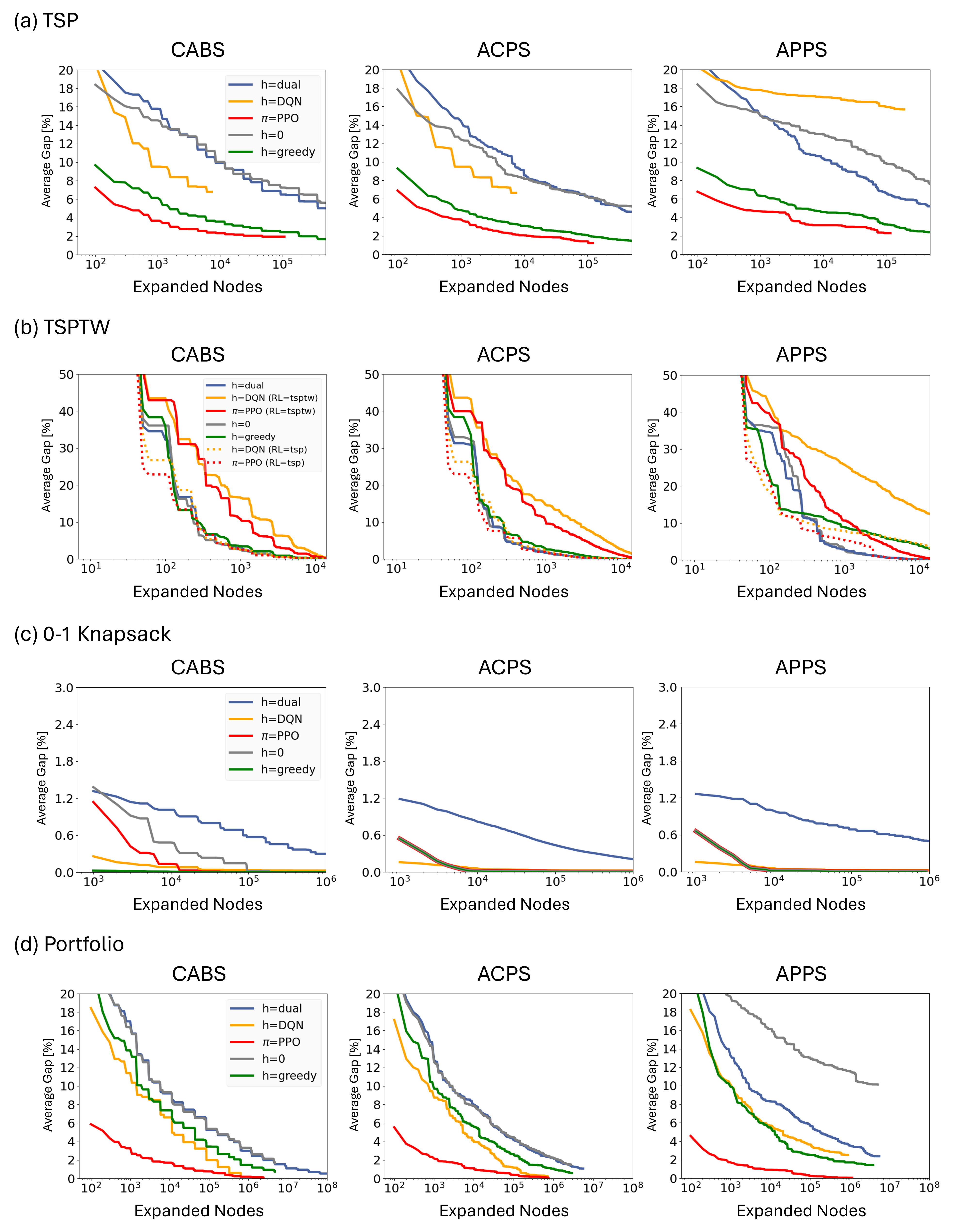}
    \end{center}
    \caption{Results of applying heuristics to guide DIDP, averaged over 40 instances (20 each for small and medium sizes). Small instances have $n=20$ and medium instances have $n=50$, except for 0-1 Knapsack ($n=50$ small, $n=100$ medium).} 
    \label{fig:result_node_expansion}
\end{figure}

Figure~\ref{fig:result_node_expansion} shows the solution quality per node expansion with different heuristic guidance for each problem and DIDP search algorithm. 

\vspace{0.1cm}
\noindent \textbf{TSP }
The plots highlight the strong performance of PPO guidance (red) compared to other heuristics, including dual-bound (blue) and greedy heuristic (gre-en), across all three solvers. DQN also outperforms the default dual-bound guidance except for APPS, though it falls short of 
the problem-specific greedy heuristic. Greedy heuristic guidance significantly outperformed dual-bound guidance.



\vspace{0.1cm}
\noindent \textbf{TSPTW }
The solid yellow and red lines~(denoted as ``RL$=$tsptw'') represent the performance of DIDP guided by the RL agent trained in the TSPTW environment. The performance of these DIDP was significantly worse than that of dual-bound and greedy heuristic guidance. In fact, their performance was even worse than $h=0$. 
Given these results, we experimented with using the TSP RL model to guide the search~(dotted lines), which significantly outperformed DIDP guided by the TSPTW RL model but achieved about the same performance as other heuristic guidance methods, including $h=0$. 

\vspace{0.1cm}
\noindent \textbf{0-1 Knapsack }
All the heuristic guidance quickly achieved solutions with gaps of less than 1\%, although dual-bound guidance (blue) exhibited slightly worse performance compared to the others. During training, the policies generated by DQN and PPO rapidly converged to the best-ratio heuristic~(greedy heuristic), which explains the similar behavior across these three guidance methods.

\vspace{0.1cm}
\noindent \textbf{Portfolio Optimization}
The plots highlight that PPO guidance~(red) significantly outperforms other guidance, including the problem-specific greedy guidance~(green). DQN guidance~(yellow) also surpassed the dual-bound guidance~(blue) and was competitive with the greedy heuristic.

\smallskip
Table~\ref{table:result_baselines} shows the performance of the different methods. DIDP performed best in TSPTW, MIP~(Gurobi) in TSP and Knapsack, and CP Optimizer for Portfolio. DIDP guided by PPO outperforms the dual-bound guidance using the same solver in terms of average gap across all problem domains except for TSP with $n=50$. As CABS with dual-bound guidance is the best performing solver~\cite{kuroiwa2024journal}, these results suggest that PPO guidance has surpassed the current state-of-the-art for DIDP in these problems. DIDP guided by DQN also outperformed dual-bound in several settings, though it falls short in TSP. 

RL guidance takes orders of magnitude more time for per node expansion due to the call to the neural network prediction. Despite this bottleneck, PPO guidance achieves better performance than the baselines at the time limit. Compared to Cappart et al.~\cite{cappart2021combining}, the DQN-guided approach in our framework achieves significantly higher performance. For instance, in Portfolio, BaB-DQN achieves an average gap of 10.8\%, while CABS ($h$=DQN) achieves a substantially lower gap of 0.77\%. Similarly, DIDP with $h$=DQN outperforms BaB-DQN in TSPTW, likely because the base DIDP model substantially outperforms the base CP model. PPO guidance exhibits a similar trend, showing notable improvements over RBS-PPO in TSPTW and slightly better results in Portfolio (e.g., 0.50\% for RBS-PPO compared to 0.19\% for CABS ($\pi$=PPO)). However, in TSP, RBS-PPO shows slightly better performance than DIDP guided by PPO.

Table~\ref{table:result_baselines} also compares the performance of heuristic sampling against baseline methods. In TSP, PPO clearly outperformed other heuristics, achieving average gaps of 0.26\% for $n=20$ and 3.85\% for $n=50$. In TSPTW, DQN and PPO heuristics were relatively effective in finding feasible solutions~(e.g., achieving feasibility in 16 out of 20 instances for $n=20$), but their ability in optimizing the solution cost is poor (average gaps of 35.12\% for DQN and 25.64\% for PPO for $n=20$). For Knapsack, the performance across heuristics was similar, although the dual-bound heuristic was slightly worse. In Portfolio, PPO showed strong standalone performance, being only 0.75\% worse than the best-known solution for $n=50$. The average gap for DQN~(3.63\% for $n=20$ and 8.09\% for $n=50$) is comparable to that of the greedy heuristic~(5.22\% for $n=20$ and 6.84\% for $n=50$), while the dual-bound heuristic performed considerably worse~(4.19\% for $n=20$ and 18.46\% for $n=50$).

\begin{table}[tb]
\caption{Comparison of results with baseline methods. Values represent averages over 20 instances. The lowest average gap for each problem and size $n$ is underlined. The symbol * indicates that optimality was proven for all 20 instances. In the DIDP results, values are highlighted in bold if the corresponding method achieves a better average gap than dual-bound guidance using the same solver. 
For TSP with $n=50$, sampling with DQN timed-out before completing 1280 samples. 
``-'' denotes reaching either time or memory limits. ``t.o.'' indicates that all 20 instances reached the time limit.}
\resizebox{\columnwidth}{!}{%
\begin{tabular}{lll|rrrr|rrrrrr|rrrr|rrrr}
\rowcolor{black}
                                                 &                                            &          & \multicolumn{4}{c|}{\color{white} TSP}                                                                                  & \multicolumn{6}{c|}{\color{white} TSPTW}                                                                                                                                                & \multicolumn{4}{c|}{\color{white} 0-1 Knapsack}                                                                                 & \multicolumn{4}{c}{\color{white} Portfolio}                                                                                    \\
\rowcolor{black}\multicolumn{3}{l|}{Method}                                                                              & \multicolumn{2}{c}{\color{white} n=20}                            & \multicolumn{2}{c|}{\color{white} n=50}                           & \multicolumn{3}{c}{\color{white} n=20}                                                                & \multicolumn{3}{c|}{\color{white} n=50}                                                       & \multicolumn{2}{c}{\color{white} n-50}                                    & \multicolumn{2}{c|}{\color{white} n=100}                          & \multicolumn{2}{c}{\color{white} n=20}                                    & \multicolumn{2}{c}{\color{white} n=50}                           \\ \hline
\multicolumn{1}{l|}{Type}                        & \multicolumn{2}{l|}{Name}                             & \multicolumn{1}{c}{Gap} & \multicolumn{1}{c|}{Time} & \multicolumn{1}{c}{Gap} & \multicolumn{1}{c|}{Time} & \multicolumn{1}{c}{Feas.} & \multicolumn{1}{c}{Gap} & \multicolumn{1}{c|}{Time}         & \multicolumn{1}{c}{Feas.} & \multicolumn{1}{c}{Gap} & \multicolumn{1}{c|}{Time} & \multicolumn{1}{c}{Gap} & \multicolumn{1}{c|}{Time}         & \multicolumn{1}{c}{Gap} & \multicolumn{1}{c|}{Time} & \multicolumn{1}{c}{Gap} & \multicolumn{1}{c|}{Time}         & \multicolumn{1}{c}{Gap} & \multicolumn{1}{c}{Time} \\ \hline
\multicolumn{1}{l|}{\multirow{3}{*}{CP}}         & \multicolumn{2}{l|}{CP Optimizer}                     & 0.00                    & \multicolumn{1}{r|}{25}   & 2.75                    & t.o.                      & 20                        & 12.04                   & \multicolumn{1}{r|}{t.o.}         & 20                        & 32.32                   & t.o.                      & 0.00                    & \multicolumn{1}{r|}{t.o.}         & 0.02                    & t.o.                      & \underline{0.00*}                   & \multicolumn{1}{r|}{\textless{}1} & \underline{0.00}                    & t.o.                     \\
\multicolumn{1}{l|}{}                            & \multicolumn{2}{l|}{BaB-DQN}                          & 3.77                    & \multicolumn{1}{r|}{t.o.} & 18.46                   & t.o.                      & 20                        & \underline{0.00*}                   & \multicolumn{1}{r|}{216}          & 20                        & 40.83                   & t.o.                      & 0.00                    & \multicolumn{1}{r|}{t.o.}         & 0.00                    & t.o.                      & \underline{0.00*}                   & \multicolumn{1}{r|}{1270}         & 10.18                   & t.o.                     \\
\multicolumn{1}{l|}{}                            & \multicolumn{2}{l|}{RBS-PPO}                          & 0.12                   & \multicolumn{1}{r|}{t.o.} & 1.17                  & t.o.                      & 20                        & 0.65                    & \multicolumn{1}{r|}{t.o.}         & 20                        & 33.82                   & t.o.                      & 0.00                    & \multicolumn{1}{r|}{t.o.}         & 0.00                    & t.o.                      & \underline{0.00*}                   & \multicolumn{1}{r|}{487}          & 0.50                    & t.o.                     \\ \hline
\multicolumn{1}{l|}{MIP}                         & \multicolumn{2}{l|}{Gurobi}                           & \underline{0.00*}                   & \multicolumn{1}{r|}{3}    & \underline{0.03}                    & t.o.                      & 20                        & \underline{0.00*}                   & \multicolumn{1}{r|}{\textless{}1} & 20                        & \underline{0.00*}                   & 119                       & \underline{0.00*}                   & \multicolumn{1}{r|}{\textless{}1} & \underline{0.00*}                   & \textless{}1              & -                       & \multicolumn{1}{r|}{-}            & -                       & -                        \\ \hline
\multicolumn{1}{l|}{\multirow{4}{*}{Heuristics}} & \multicolumn{2}{l|}{DQN}                              & 2.29                    & \multicolumn{1}{r|}{1251} & (20.14)                 & (18239)                   & 16                        & 35.12                   & \multicolumn{1}{r|}{1405}         & 0                         & -                       & -                         & 0.02                    & \multicolumn{1}{r|}{44}           & 0.04                    & 121                       & 3.63                    & \multicolumn{1}{r|}{36}           & 8.09                    & 132                      \\
\multicolumn{1}{l|}{}                            & \multicolumn{2}{l|}{PPO}                              & 0.26                    & \multicolumn{1}{r|}{249}  & 3.85                    & 1783                      & 20                        & 25.64                   & \multicolumn{1}{r|}{423}          & 20                        & 46.93                   & 2050                      & 0.04                    & \multicolumn{1}{r|}{51}           & 0.13                    & 132                       & 3.35                    & \multicolumn{1}{r|}{38}           & 0.75                    & 111                      \\
\multicolumn{1}{l|}{}                            & \multicolumn{2}{l|}{Dual-bounds}                      & 2.91                    & \multicolumn{1}{r|}{4}    & 9.58                    & 1424                      & 0                         & -                       & \multicolumn{1}{r|}{-}            & 0                         & -                       & -                         & 0.07                    & \multicolumn{1}{r|}{8}            & 0.37                    & 44                        & 4.19                    & \multicolumn{1}{r|}{1}            & 18.46                   & 4                        \\
\multicolumn{1}{l|}{}                            & \multicolumn{2}{l|}{Greedy}                           & 5.30                    & \multicolumn{1}{r|}{11}   & 8.48                    & 176                       & 0                         & -                       & \multicolumn{1}{r|}{-}            & 0                         & -                       & -                         & 0.03                    & \multicolumn{1}{r|}{3}            & 0.01                    & 8                         & 5.22                    & \multicolumn{1}{r|}{2}            & 6.84                    & 7                        \\ \hline
\multicolumn{1}{l|}{\multirow{12}{*}{DIDP}}      & \multicolumn{1}{l|}{\multirow{4}{*}{CABS}} & $h$=greedy & 0.00                    & \multicolumn{1}{r|}{292}  & 2.79                    & -                         & 20                        & \underline{0.00*}                   & \multicolumn{1}{r|}{\textless{}1} & 20                        & \underline{0.00*}                   & 22                        & 0.00                    & \multicolumn{1}{r|}{-}            & \textbf{0.00}                    & -                         & \underline{0.00*}                   & \multicolumn{1}{r|}{15}           & 1.48                    & -                        \\
\multicolumn{1}{l|}{}                            & \multicolumn{1}{l|}{}                      & $h$=dual   & \underline{0.00*}                   & \multicolumn{1}{r|}{19}   & 1.86                    & -                         & 20                        & \underline{0.00*}                   & \multicolumn{1}{r|}{\textless{}1} & 20                        & \underline{0.00*}                   & \textless{}1              & 0.00                    & \multicolumn{1}{r|}{-}            & 0.09                    & -                         & \underline{0.00*}                   & \multicolumn{1}{r|}{2}            & 0.81                    & -                        \\
\multicolumn{1}{l|}{}                            & \multicolumn{1}{l|}{}                      & $h$=DQN    & 0.00                    & \multicolumn{1}{r|}{154}  & 12.05                   & t.o.                      & 20                        & \underline{0.00*}                   & \multicolumn{1}{r|}{44}           & 20                        & \underline{0.00*}                   & 765                       & 0.00                    & \multicolumn{1}{r|}{t.o.}         & \textbf{0.05}                    & t.o.                      & \underline{0.00*}                   & \multicolumn{1}{r|}{780}          & \textbf{0.77}                    & t.o.                     \\
\multicolumn{1}{l|}{}                            & \multicolumn{1}{l|}{}                      & $\pi$=PPO   & 0.00                    & \multicolumn{1}{r|}{32}   & 3.89                    & t.o.                      & 20                        & \underline{0.00*}                   & \multicolumn{1}{r|}{40}           & 20                        & \underline{0.00*}                   & 860                       & 0.00                    & \multicolumn{1}{r|}{t.o.}         & \textbf{0.00}                    & t.o.                      & \underline{0.00*}                   & \multicolumn{1}{r|}{477}          & \textbf{0.19}                    & t.o.                     \\ \cline{2-21} 
\multicolumn{1}{l|}{}                            & \multicolumn{1}{l|}{\multirow{4}{*}{ACPS}} & $h$=greedy & \underline{0.00*}                   & \multicolumn{1}{r|}{62}   & \textbf{2.45}                    & -                         & 20                        & \underline{0.00*}                   & \multicolumn{1}{r|}{\textless{}1} & 20                        & \underline{0.00*}                   & 2                         & 0.00                    & \multicolumn{1}{r|}{-}            & \textbf{0.00}                    & -                         & \underline{0.00*}                   & \multicolumn{1}{r|}{4}            & \textbf{1.17}                    & -                        \\
\multicolumn{1}{l|}{}                            & \multicolumn{1}{l|}{}                      & $h$=dual   & \underline{0.00*}                   & \multicolumn{1}{r|}{8}    & 6.35                    & -                         & 20                        & \underline{0.00*}                   & \multicolumn{1}{r|}{\textless{}1} & 20                        & \underline{0.00*}                   & \textless{}1              & 0.00                    & \multicolumn{1}{r|}{-}            & 0.21                    & -                         & \underline{0.00*}                   & \multicolumn{1}{r|}{\textless{}1} & 2.14                    & -                        \\
\multicolumn{1}{l|}{}                            & \multicolumn{1}{l|}{}                      & $h$=DQN    & 0.00                    & \multicolumn{1}{r|}{371}  & 10.09                   & t.o.                      & 20                        & \underline{0.00*}                   & \multicolumn{1}{r|}{11}           & 20                        & \underline{0.00*}                   & 99                        & 0.00                    & \multicolumn{1}{r|}{t.o.}         & \textbf{0.03}                    & t.o.                      & \underline{0.00*}                   & \multicolumn{1}{r|}{180}          & \textbf{0.50}                    & t.o.                     \\
\multicolumn{1}{l|}{}                            & \multicolumn{1}{l|}{}                      & $\pi$=PPO   & 0.00                    & \multicolumn{1}{r|}{345}  & \textbf{2.45}                    & t.o.                      & 20                        & \underline{0.00*}                   & \multicolumn{1}{r|}{11}           & 20                        & \underline{0.00*}                   & 123                       & 0.00                    & \multicolumn{1}{r|}{t.o.}         & \textbf{0.00}                    & t.o.                      & \underline{0.00*}                   & \multicolumn{1}{r|}{119}          & \textbf{0.14}                    & t.o.                     \\ \cline{2-21} 
\multicolumn{1}{l|}{}                            & \multicolumn{1}{l|}{\multirow{4}{*}{APPS}} & $h$=greedy & \underline{0.00*}                   & \multicolumn{1}{r|}{66}   & \textbf{3.46}                    & -                         & 20                        & \underline{0.00*}                   & \multicolumn{1}{r|}{\textless{}1} & 20                        & \underline{0.00*}                   & 4                         & 0.00                    & \multicolumn{1}{r|}{-}            & \textbf{0.00}                    & -                         & \underline{0.00*}                   & \multicolumn{1}{r|}{4}            & \textbf{2.86}                    & -                        \\
\multicolumn{1}{l|}{}                            & \multicolumn{1}{l|}{}                      & $h$=dual   & \underline{0.00*}                   & \multicolumn{1}{r|}{8}    & 8.30                    & -                         & 20                        & \underline{0.00*}                   & \multicolumn{1}{r|}{\textless{}1} & 20                        & \underline{0.00*}                   & \textless{}1              & 0.00                    & \multicolumn{1}{r|}{-}            & 0.58                    & -                         & \underline{0.00*}                   & \multicolumn{1}{r|}{\textless{}1} & 4.74                    & -                        \\
\multicolumn{1}{l|}{}                            & \multicolumn{1}{l|}{}                      & $h$=DQN    & 1.08                    & \multicolumn{1}{r|}{t.o.} & 31.57                   & t.o.                      & 20                        & \underline{0.00*}                   & \multicolumn{1}{r|}{13}           & 20                        & \underline{0.00*}                   & 141                       & 0.00                    & \multicolumn{1}{r|}{t.o.}         & \textbf{0.03}                    & t.o.                      & \underline{0.00*}                   & \multicolumn{1}{r|}{187}          & 5.01                    & t.o.                     \\
\multicolumn{1}{l|}{}                            & \multicolumn{1}{l|}{}                      & $\pi$=PPO   & 0.20                    & \multicolumn{1}{r|}{t.o.} & \textbf{4.17}                    & t.o.                      & 20                        & \underline{0.00*}                   & \multicolumn{1}{r|}{12}           & 20                        & \underline{0.00*}                   & 141                       & 0.00                    & \multicolumn{1}{r|}{t.o.}         & \textbf{0.00}                    & t.o.                      & \underline{0.00*}                   & \multicolumn{1}{r|}{119}          & \textbf{0.10}                    & t.o.                    
\end{tabular}%
}
\label{table:result_baselines}
\end{table}

\section{Discussion}
\noindent \textbf{When is RL guidance helpful, and to what extent? } RL guidance is most impactful when heuristic quality plays a critical role in solution quality. For instance, in TSPTW, the DIDP model prunes many states by time windows. In such cases, the primary role of heuristic guidance in DIDP appears to be minimizing solution costs rather than ensuring feasibility. In contrast, TSP and Portfolio DP models lack such pruning mechanisms, making heuristic quality a more critical factor in improving the solution quality. 

When the performance of DIDP appears to depend primarily on heuristic guidance, the effectiveness of guidance aligns with the performance of the heuristic in sampling-based approaches. For example, PPO guidance consistently outperforms dual-bound guidance because the PPO heuristic is better at driving the search towards high quality solution, as shown in Table~\ref{table:result_baselines}. Dual-bounds are admissible and thus effective at de-prioritizing unpromising decisions, but may not necessarily guide the search towards more promising solutions.



\noindent \textbf{Solution Quality in Terms of Solve Time }
While DQN and PPO guidance demonstrate significantly higher solution quality per node expansion compared to dual-bound guidance, 
their performance gains over time are relatively limited. The primary cause lies in the time required to expand a single node. As shown in Table~\ref{table:result_baselines}, generating a solution using DQN or PPO takes much longer than using dual-bound or greedy heuristics~(e.g., DQN takes 313 times longer than dual-bound to sample 1280 times for TSP $n$=50). While our experiments highlight the potential of RL-based heuristics, they also emphasize the need to address the computational overhead associated with these methods. 


\section{Conclusion}
The initial demonstration of DIDP solver performance was based on search guidance with dual bounds defined in the model. Through experiments on three anytime algorithms, we demonstrated that RL can provide heuristic guidance that improves solution quality with fewer node expansions. These findings show the effectiveness of RL-guided search within anytime algorithms and help to elucidate the conditions where RL guidance is most beneficial, such as in domains where heuristic quality plays a critical role in solution improvement. The inherent structural similarity between DP and RL models offers a natural synergy, enabling RL to be easily integrated into the DIDP framework. With further work on automating RL model building and reducing the time to evaluate states, RL-guided DIDP has the potential to serve as a practical and powerful tool for combinatorial optimization.

%
%
%
\bibliographystyle{splncs04}

%





\appendix

\section{Search Algorithms}\label{appendix:didp_search}
A solution of a DyPDL model can be computed by a cost-algebraic search algorithm~\cite{caasdy}. We briefly describe the three search algorithms used in the experiments. A full description of these algorithms is found in Kuroiwa and Beck~\cite{kuroiwa2024journal}.

\subsubsection{Complete Anytime Beam Search (CABS)}
Beam search is a heuristic breadth-first search algorithm that maintains only the best $b$ states at each depth. While beam search is not a complete algorithm, CABS~\cite{zhang1998complete} ensures optimality by running multiple iterations of beam search, doubling the beam width after each iteration. 

The algorithm maintains a set of states to expand at the current depth in an open list $O$. Starting from $O=\{s_0\}$, beam search expands all states in $O$, generating a successor state $s'=s[\![\tau ]\!]$ for each state, $s \in O$, and transition $\tau$ applicable in $s$. Each successor state $s'$ must satisfy all state constraints $\mathcal{C}$, i.e., $s'\models \mathcal{C}$. At each successor state, the $f$-value is computed as $f(s')=g(s')+h(s')$. $s'$ is \textit{pruned}~(not added to $O$) if $g(s') + \eta(s')\geq\bar{\zeta}$, where $\bar{\zeta}$ is the incumbent solution cost~($g(s') + \eta(s')\leq\bar{\zeta}$ for maximization problems). After pruning, the best $b$ states based on the $f$-values are retained in $O$ and the process continues. If $s$ is a base state, then the beam search terminates; the path from $s_0$ to $s$ is a solution, and the incumbent solution is updated if $g(s) + \min_{\{B| B\in \mathcal{B}; s\models \mathcal{C}_B\}}$\text{\fontfamily{cmss}\selectfont cost}$_B(s)<\bar{\zeta}$~(for maximization, $g(s) + \max_{\{B| B\in \mathcal{B}; s\models \mathcal{C}_B\}}$\text{\fontfamily{cmss}\selectfont cost}$_B(s) >\bar{\zeta}$). The beam width $b$ is then doubled, and the algorithm repeats. CABS terminates when optimality is proven. In the default DIDP implementation, the dual-bound function $\eta$ is used as $h$.

\subsubsection{Anytime Column Progressive Search (ACPS)}
ACPS~\cite{acs} partitions the open list $O$ into layers $O_i$ for each depth $i$. States are grouped into the same layer if they have the same number of transitions from the target state $s_0$. Starting from $i=0$, $O_0=\{s_0\}$ and $O_i=\emptyset$ for $i>0$, ACPS expands best $b$ states in $O_i$ based on $f$-values, inserts successor states into $O_{i+1}$, and increments $i$ by 1. ACPS starts from $b=1$ and increases $b$ by 1 when it reaches the maximum depth. $i$ is reset to $0$ when a new best solution is found or $O_i=\emptyset$ for all $j\geq i$.

\subsubsection{Anytime Pack Progressive Search (APPS)}
ACPS~\cite{aps} maintains three sets: the best states $O_{best}\subseteq O$, the best successor states $O_c\subseteq O$, and a suspend list $O_s\subseteq O$. Starting from $O_{best}=\{s_0\}$, $O_c=\emptyset$, and $O_s=\emptyset$, APPS expands the states from $O_{best}$ and inserts the best $b$ successor states into $O_c$ and other successor states in $O_s$. Then, $O_{best}$ and $O_c$ are swapped and the procedure is continued. When $O_{best}$ and $O_c$ are empty, the best $b$ states from $O_s$ are moved to $O_{best}$. Following the implementation by Kuroiwa and Beck~\cite{kuroiwa2024journal}, the algorithm starts with $b=1$ and increases $b$ by 1 after the best $b$ states are moved from $O_s$ to $O_{best}$.

\section{Network Architecture}\label{appendix:architecture}

\subsection{TSP}
TSP instances are naturally represented as fully connected graphs, where each vertex corresponds to a customer and edges represent the distances between customers. To leverage this graph structure, we used a graph attention network (GAT)~\cite{gat} for the state representation. The network architecture was inspired by the implementation from Cappart et al.~\cite{cappart2021combining}. Static features are the x-y coordinates of the customers and the dynamic features correspond to the state variables mapped from the DIDP model: a binary feature that indicates if the customer is in the unvisited set $U$ and second binary feature indicating whether the customer is the last visited location ($i$ in the DIDP model). For DQN, the dimension of the last layer corresponds to the action set size, i.e., $|A|$, and the output values are the estimation of the Q-values for the actions. For PPO, two separate networks are used, an actor network and a critic network, both sharing the same GAT architecture for their input layers. The critic network’s final layer outputs a single value, representing the estimated value of the state, while the actor network applies a softmax activation function to its final layer to output action probabilities. For both models, the outputs are masked to exclude actions that are not applicable in the current state. The hyperparameters used for this RL model are summarized in Table~\ref{table:hyperparams}.

\begin{table}[tb]
\caption{Hyperparameters values used in the RL training.}
\resizebox{\columnwidth}{!}{%
\begin{tabular}{l|rrrr|rrrr|rrrr|rrrr}
\rowcolor{black}\multicolumn{1}{c|}{\color{white} Parameter} & \multicolumn{4}{c|}{\color{white} TSP}                                                                                   & \multicolumn{4}{c|}{\color{white} TSPTW}                                                                                 & \multicolumn{4}{c|}{\color{white} 0-1 Knapsack}                                                                            & \multicolumn{4}{c}{\color{white} Portfolio}                                                                             \\
\rowcolor{black}                               & \multicolumn{2}{c}{\color{white} DQN}                             & \multicolumn{2}{c|}{\color{white} PPO}                             & \multicolumn{2}{c}{\color{white} DQN}                             & \multicolumn{2}{c|}{\color{white} PPO}                             & \multicolumn{2}{c}{\color{white} DQN}                              & \multicolumn{2}{c|}{\color{white} PPO}                              & \multicolumn{2}{c}{\color{white} DQN}                             & \multicolumn{2}{c}{\color{white} PPO}                             \\
\rowcolor{black}                               & \multicolumn{1}{c}{\color{white} n=20} & \multicolumn{1}{c}{\color{white} n=50} & \multicolumn{1}{c}{\color{white} n=20} & \multicolumn{1}{c|}{\color{white} n=50} & \multicolumn{1}{c}{\color{white} n=20} & \multicolumn{1}{c}{\color{white} n=50} & \multicolumn{1}{c}{n=20} & \multicolumn{1}{c|}{\color{white} n=50} & \multicolumn{1}{c}{\color{white} n=50} & \multicolumn{1}{c}{\color{white} n=100} & \multicolumn{1}{c}{\color{white} n=50} & \multicolumn{1}{c|}{\color{white} n=100} & \multicolumn{1}{c}{\color{white} n=20} & \multicolumn{1}{c}{\color{white} n=50} & \multicolumn{1}{c}{\color{white} n=20} & \multicolumn{1}{c}{\color{white} n=50} \\ \hline
Batch size                     & 128                      & 256                      & 256                      & 256                       & 32                       & 64                       & 128                      & 64                        & 128                      & 128                       & 128                      & 128                        & 64                       & 128                      & 128                      & 128                      \\
Learning rate                  & $10^{-4}$                & $10^{-4}$                & $10^{-4}$                & $10^{-4}$                 & $10^{-4}$                & $10^{-4}$                & $10^{-4}$                & $10^{-4}$                 & $10^{-4}$                & $10^{-4}$                 & $10^{-3}$                & $10^{-3}$                  & $10^{-5}$                & $10^{-5}$                & $10^{-5}$                & $10^{-5}$                \\
\# GAT layers                  & 4                        & 4                        & 4                        & 4                         & 4                        & 4                        & 4                        & 4                         & -                        & -                         & -                        & -                          & -                        & -                        & -                        & -                        \\
Embedding dimension            & 64                       & 64                       & 128                      & 128                       & 32                       & 64                       & 256                      & 128                       & -                        & -                         & -                        & -                          & 40                       & 40                       & 40                       & 40                       \\
\# Hidden layers               & 3                        & 3                        & 4                        & 4                         & 2                        & 3                        & 4                        & 4                         & 2                        & 3                         & 3                        & 3                          & 2                        & 3                        & 2                        & 2                        \\
Hidden layer dimension         & 64                       & 64                       & 128                      & 128                       & 32                       & 64                       & 256                      & 128                       & 128                      & 128                       & 128                      & 128                        & 128                      & 256                      & 128                      & 128                      \\
Reward scaling factor          & $10^{-3}$                & $10^{-3}$                & $10^{-3}$                & $10^{-3}$                 & $10^{-3}$                & $10^{-3}$                & $10^{-3}$                & $10^{-3}$                 & $10^{-4}$                & $10^{-4}$                 & $10^{-4}$                & $10^{-4}$                  & $10^{-3}$                & $10^{-3}$                & $10^{-3}$                & $10^{-3}$                \\ \hline
Softmax temperature            & 10                       & 2                        & -                        & -                         & 10                       & 10                       & -                        & -                         & 2                        & 2                         & -                        & -                          & 10                       & 10                       & -                        & -                        \\ \hline
Entropy value                  & -                        & -                        & $10^{-3}$                & $10^{-3}$                 & -                        & -                        & $10^{-3}$                & $10^{-3}$                 & -                        & -                         & $10^{-3}$                & $10^{-3}$                  & -                        & -                        & $10^{-3}$                & $10^{-3}$                \\
Clipping value                 & -                        & -                        & 0.1                      & 0.1                       & -                        & -                        & 0.1                      & 0.1                       & -                        & -                         & 0.1                      & 0.1                        & -                        & -                        & 0.1                      & 0.1                      \\
\# Epochs per update           & -                        & -                        & 3                        & 3                         & -                        & -                        & 3                        & 3                         & -                        & -                         & 4                        & 4                          & -                        & -                        & 4                        & 4                       
\end{tabular}%
}
\label{table:hyperparams}
\end{table}

\subsection{TSPTW}
Similarly to TSP, both DQN and PPO use a GAT as the input representation, and has the same output format. Each node in GAT has two additional static features to encode the time windows $[a_j, b_j]$ for customer $j\in N$. The hyperparameters used for this RL model are summarized in Table~\ref{table:hyperparams}.

\subsection{0-1 Knapsack}
We used Deep Sets~\cite{deepsets} for state representation, adopting the implementation from the Point-Cloud Classification model by Zaheer et al.~\cite{deepsets}. The architecture first applies a function $\phi$ to each element in the set independently, and then these outputs are aggregated using a permutation-invariant function, such as sum, mean, and max~(mean is used in our implementation). The network representing $\phi$ consists of three permutation equivariant layers with 256 channels followed by a max-pooling layer. The resulting vector representation is then input to a fully-connected layer, followed by a tanh activation function. Each item has four static features and four dynamic features. The static features are the weight $w_j$, profit $p_j$, the profit-weight ratio $w_j/p_j$, and the inverse profit-weight ratio $p_j/w_j$ of the item $j$. The dynamic features are the remaining capacity after taking item $j$, a binary feature indicating whether $j$ is yet to be considered (i.e., 1 if $j > i$ for the current item index $i$ in the DIDP model), a binary feature indicating if $j$ is the current item being considered (i.e., 1 if $j=i$), and a binary feature indicating whether the capacity is exceeded if we select item $j$. The hyperparameters used for this RL model are summarized in Table~\ref{table:hyperparams}.

\subsection{Portfolio Optimization}
For this problem domain, we used a permutation-invariant set transformer~\cite{settransformer} as the state representation. Similarly to GAT, the set transformer produces an embedding for each investment to form an embedding layer, which is then followed by fully-connected layers. The number of fuly-connected layers is 3 for DQN $n=50$ and 2 for others~(see Table~\ref{table:hyperparams}). Each investment has five static features and four dynamic features. The static features are the weight $w_j$, mean $\mu_j$, deviation $\sigma_j$, skewness $\gamma_j$, and kurtosis $\kappa_j$ of item $j$ (each parameter is normalized by the L2 norm of all the investments for the instance). The dynamic features are the L2-normalized remaining capacity after selecting investment $j$, a binary feature indicating if investment $j$ is already considered, another binary feature indicating if this investment is the current investment being considered ($i$ in the DIDP model), and a binary feature indicating if the budget will be exceeded if we take investment $j$. The hyperparameters used for this RL model are summarized in Table~\ref{table:hyperparams}.

\section{Greedy heuristics}\label{appendix:greedy}
In greedy heuristics, the $h$-values at a state are the cost of the path from that state to a base case found by rolling-out successor states according to the greedy heuristic. For TSP, starting from the target state~($\langle U, i, t\rangle =\langle\{0, ..., n\}, 0, 0\rangle$), an unvisited customer with minimum distance from the current customer $i$, $\text{argmin}_{j\in U} c_{ij}$, is iteratively chosen, until $U$ becomes empty. For TSPTW, an unvisited customer with the minimum $\max(t + c_{ij}, a_j)$ is chosen instead and the time window constraints are otherwise ignored. 
 For 0-1 Knapsack, items are added in the descending order of profit-weight ratio $(p_j/w_j)$ until the capacity limit is reached. For Portfolio Optimization, items are considered in the descending order of efficiency $(\lambda_1 \mu_j-\lambda_2\sigma_j+\lambda_3\gamma_h-\lambda_4\kappa_j)/w_j$ and added if the capacity limit is not exceeded.

\section{Dual bounds for Portfolio Optimization}\label{appendix:portfolio_duals}
We introduced two dual bounds for the Portfolio Optimization problem.

\subsection{Sum of the remaining items}

The first dual bound is derived by considering only the positive terms in the cost function (i.e., mean and skewness). The dual bound is the cost function for the remaining items, assuming all the remaining items are taken, considering only mean and skewness:

\begin{equation}
    V(x, i, Y)\leq \lambda_1 \sum_{j=i}^{n-1}\mu_j+\lambda_3\sqrt[3]{\sum_{j=i}^{n-1}\gamma^3_j}\label{eq:apdx_d1_1}
\end{equation}

\noindent where $i$ is the current item index. 

\textit{Proof.} We evaluate the validity of the dual bound term-wise. 
Suppose we make optimal transitions from a given state $s=(x, i, Y)$ to the goal state such that it maximizes $V(x, i, Y)$. We denote such optimal investments from stage $i$ $\{y^*_i...y^*_{n-1}\}$. Then,
\begin{equation}
    h^*_{means}=\lambda_1 \sum_{j=i}^{n-1}\mu_j y^*_j
\end{equation}
is the means term of the optimal cost at a given state, following $\{y^*_i...y^*_{n-1}\}$. The first term in Eq.~(\ref{eq:apdx_d1_1}) will never underestimate $h^*_{means}$:
\begin{align}
    h^*_{means} = \lambda_1 \sum_{j=i}^{n-1}\mu_j y^*_j \leq \lambda_1 \sum_{j=i}^{n-1}\mu_j
\end{align}

\noindent since $y^*_j$ is binary for all $j\in \{i...n-1\}$.



Let $h^*_{skew}$ denote the skewness term of the optimal cost at a given state at stage $i$, and $g_{skew}(=\lambda_3\sqrt[3]{\sum_{j=0}^{i-1}\gamma^3_j y_j})$ the skewness term of the path cost to this state. Then, the optimal $f$-value at state $s$, $g_{skew}+h^*_{skew}$, corresponds to the skewness term of the solution cost obtained by following the decisions $y_0...y_{i-i}$, up to stage $i$, and then taking the optimal decisions $y^*_{i}...y^*_{n-1}$ thereafter. Hence,
\begin{equation}
    g_{skew}+h^*_{skew} = \lambda_3\sqrt[3]{\sum_{j=0}^{i-1}\gamma^3_j y_j + \sum_{j=i}^{n-1}\gamma^3_j y^*_j}.
\end{equation}



Since the sum of cube roots is always greater than or equal to the cube root of all the elements,

\begin{align}
    g_{skew}+h^*_{skew} &\leq \lambda_3\sqrt[3]{\sum_{j=0}^{i-1}\gamma^3_j y_j + \sum_{j=i}^{n-1}\gamma^3_j y^*_j}\\
    &\leq \lambda_3\sqrt[3]{\sum_{j=0}^{i-1}\gamma^3_j y_j} + \lambda_3\sqrt[3]{\sum_{j=i}^{n-1}\gamma^3_j y^*_j} \\
    &\leq \lambda_3\sqrt[3]{\sum_{j=0}^{i-1}\gamma^3_j y_j} + \lambda_3\sqrt[3]{\sum_{j=i}^{n-1}\gamma^3_j}
\end{align}

Since $g_{skew}= \lambda_3\sqrt[3]{\sum_{j=0}^{i-1}\gamma^3_j y_j}$, by subtracting $g_{skew}$ on both sides,

\begin{equation}
    h^*_{skew} \leq \lambda_3\sqrt[3]{\sum_{j=i}^{n-1}\gamma^3_j}
\end{equation}

\noindent and thus the second term also never underestimates the true cost. Therefore,
\begin{align}
    V(x, i, Y) = h^* \leq h^*_{means} + h^*_{skew} \leq \lambda_1 \sum_{j=i}^{n-1}\mu_j + \lambda_3\sqrt[3]{\sum_{i=j}^{n-1}\gamma^3_i}
\end{align}

\noindent is a valid dual bound. \qedsymbol

\subsection{Best profit-weight ratio}
The second dual bound is derived by identifying the remaining item with the best profit-weight ratio and multiplying the ratio with the remaining budget at this state:

\begin{equation}
    V(x, i, Y) \leq \max_{j\in \{i..n-1\}} \left(\frac{\lambda_1 \mu_j + \lambda_3 \sqrt[3] {\gamma^3_j}}{w_j}\right)\cdot (B-x)
\end{equation}

\noindent where $B$ is the budget at the target state and $x$ is the current total weight.

\textit{Proof.} Suppose we make optimal transitions from a given state $s=(x, i, Y)$ to the goal state such that it maximizes $V(x, i, Y)$. We denote such optimal investments from stage $i$ $\{y^*_i...y^*_{n-1}\}$. Then, the optimal $f$-value at this state $s$~(for minimization) is:

\begin{align}
    f^*_s &= \lambda_1 \left(\sum_{j=0}^{i-1}\mu_j y_j + \sum_{j=i}^{n-1}\mu_j y^*_j\right)-\lambda_2\sqrt[2]{\left(\sum_{j=0}^{i-1}\sigma^2_j y_j + \sum_{j=i}^{n-1}\sigma^2_j y^*_j\right)} \notag\\
    &\hspace{1.5cm}+ \lambda_3\sqrt[3]{\left(\sum_{j=0}^{i-1}\gamma^3_j y_j + \sum_{j=i}^{n-1}\gamma^3_j y^*_j\right)}-\lambda_4\sqrt[4]{\left(\sum_{j=0}^{i-1}\kappa^4_j y_j + \sum_{j=i}^{n-1}\kappa^4_j y^*_j\right)}\label{eq:f_star_s_1}\\
    &\leq \lambda_1 \left(\sum_{j=0}^{i-1}\mu_j y_j + \sum_{j=i}^{n-1}\mu_j y^*_j\right)- \lambda_2\sqrt[2]{\sum_{j=0}^{i-1}\sigma^2_j y_j}\notag\\
    &\hspace{1.5cm}+\lambda_3\sqrt[3]{\left(\sum_{j=0}^{i-1}\gamma^3_j y_j + \sum_{j=i}^{n-1}\gamma^3_j y^*_j\right)}-\lambda_4\sqrt[4]{\sum_{j=0}^{i-1}\kappa^4_j y_j}\label{eq:f_star_s_2}\\
    &\leq \lambda_1 \sum_{j=0}^{i-1}\mu_j y_j -\lambda_2\sqrt[2]{\sum_{j=0}^{i-1}\sigma^2_j y_j} +\lambda_3\sqrt[3]{\sum_{j=0}^{i-1}\gamma^3_j y_j} -\lambda_4\sqrt[4]{\sum_{j=0}^{i-1}\kappa^4_j}\notag\\
    &\hspace{1.5cm}+ \lambda_1 \sum_{j=i}^{n-1}\mu_j y^*_j + \lambda_3 \sqrt[3]{\sum_{j=i}^{n-1}\gamma^3_j y^*_j}\label{eq:f_star_s_3}\\
    &= g + h^*_{mean} + \overline{h^*_{skew}}
\end{align}

\noindent where $\overline{h^*_{skew}}\coloneqq\lambda_3 \sqrt[3]{\sum_{j=i}^{n-1}\gamma^3_j y^*_j}$, and $g$ is the $g$-value for state $s$. From Eq.~(\ref{eq:f_star_s_1}) to Eq.~(\ref{eq:f_star_s_2}), we ignored the second term inside the square root and the fourth root, as they only serve to decrease the right-hand side value. Then, from Eq.~(\ref{eq:f_star_s_2}) to Eq.~(\ref{eq:f_star_s_3}), we applied the fact that the sum of cube roots is always greater than the cube root of the sum of all elements. Let $h^*_s$ be the optimal $h$-value for state $s$. Since $f^*_s=g+h^*_s$, 

\begin{align}
    f^*_s&=g+h^*_s \leq g + h^*_{mean} + \overline{h^*_{skew}}\\
    \therefore \hspace{0.5cm} h^*_s& \leq h^*_{mean} + \overline{h^*_{skew}}
\end{align}

\begin{figure}[h!]
    \begin{center}
        \includegraphics[width=0.35\linewidth]{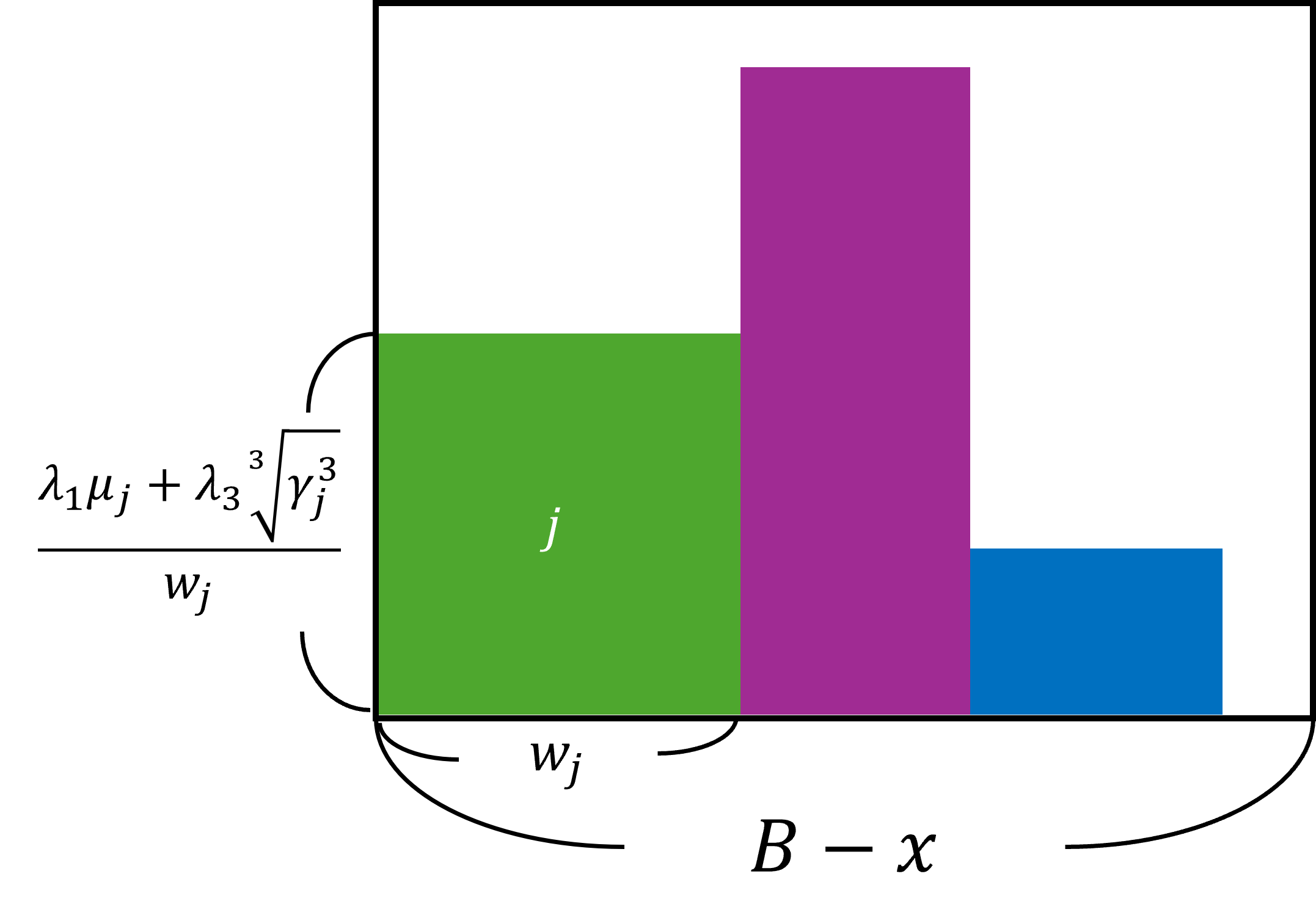}
    \end{center}
     \caption{A rectangle used to show the dual-bound based on best profit-weight ratio. Each colored box $j$ represents an investment $j$ that has not yet been considered.} 
   \label{fig:rectangle}
\end{figure}

The values of $h^*_{mean}$ and $\overline{h^*_{skew}}$ depend on the optimal decisions from stage $i$, $\{y^*_i...y^*_{n-1}\}$, which are unknown during the search. Finding an upper bound for $h^*_{mean} + \overline{h^*_{skew}}$ is equivalent to finding the set of investments that fills as much area as possible in the rectangle in Figure \ref{fig:rectangle}, where $(B-x)$ is the remaining budget at this state.

In this formulation, the maximum possible area for this rectangle is

\begin{equation}
    \max_{j\in\{i..n-1\}} \left(\frac{\lambda_1 \mu_j + \lambda_3 \sqrt[3]{\gamma^3_j}}{w_j}\right)\cdot (B-x)
\end{equation}

Therefore, 
\begin{equation}
    V(x, i, Y) = h^*_s \leq h^*_{mean} + \overline{h^*_{skew}} \leq \max_{j\in\{i..n-1\}} \left(\frac{\lambda_1 \mu_j + \lambda_3 \sqrt[3]{\gamma^3_j}}{w_j}\right)\cdot (B-x)
\end{equation}

\noindent is a valid dual bound. \qedsymbol

\end{document}